\documentclass[11pt,a4paper]{article}
\usepackage[hyperref]{eacl2021}
\usepackage{times}
\usepackage{latexsym}
\usepackage{graphicx}
\graphicspath{ {./Figures/} }

\usepackage{microtype}

\aclfinalcopy

\usepackage{mathtools}

\DeclarePairedDelimiter\floor{\lfloor}{\rfloor}

\title{A Sample-Based Training Method for Distantly Supervised Relation Extraction with Pre-Trained Transformers}

\author{Mehrdad Nasser,\textsuperscript{1}  Mohamad Bagher Sajadi,\textsuperscript{2} and Behrouz Minaei-Bidgoli\textsuperscript{1} \\
\textsuperscript{1}School of Computer Engineering, Iran University of Science and Technology\\
\textsuperscript{2}Central Tehran Branch, Islamic Azad University\\
\texttt{mehrdad\_nasser@comp.iust.ac.ir, moh.sajadi.eng@iauctb.ac.ir}\\
\texttt{b\_minaei@iust.ac.ir}

  }

\date{}

\begin{document}
\maketitle
\begin{abstract}
Multiple instance learning (MIL) has become the standard learning paradigm for distantly supervised relation extraction (DSRE). However, due to relation extraction being performed at bag level, MIL has significant hardware requirements for training when coupled with large sentence encoders such as deep transformer neural networks. In this paper, we propose a novel sampling method for DSRE that relaxes these hardware requirements. In the proposed method, we limit the number of sentences in a batch by randomly sampling sentences from the bags in the batch. However, this comes at the cost of losing valid sentences from bags. To alleviate the issues caused by random sampling, we use an ensemble of trained models for prediction. We demonstrate the effectiveness of our approach by using our proposed learning setting to fine-tuning BERT on the widely NYT dataset. Our approach significantly outperforms previous state-of-the-art methods in terms of AUC and P@N metrics. 
\end{abstract}

\section{Introduction}
Relation extraction (RE) is an essential part of Natural Language Processing (NLP) and benefits downstream tasks such as knowledge base population. The main goal of RE is to identify the semantic relationship between two entities in text. For example, based on the sentence " \textbf{\emph{Elon Musk}} is the founder of \textbf{\emph{SpaceX}}", entities \textbf{\emph{Elon Musk}} and \textbf{\emph{SpaceX}} express the \emph{founderOf} relation. Conventional supervised relation extraction methods rely on manually labeled datasets for training. The construction of such datasets requires considerable human effort and is often impractical. Distant supervision for relation extraction \cite{mintz2009distant} addresses this problem by automatically labeling entity pairs in a sentence based on their relationship in a knowledge base, removing the need for manual labeling. However, not every pair of entities in a sentence express their corresponding relation in a knowledge base; thus, distant supervision suffers from noisy labels. For example, if (\textbf{\emph{Elon Musk}}, \emph{CEOof}, \textbf{\emph{SpaceX}}) is a fact in knowledge base , distant supervision would label the aforementioned example sentence as \emph{CEOof}, which would be an incorrect label. 

Recent works have adopted multiple instance learning (MIL) framework, along with additional denoising methods to address the noisy labeling problem. In MIL, each sample in the dataset is a bag of sentences that share the same entity pair, as opposed to the conventional supervised relation extraction methods where each instance is a single sentence. Additional denoising steps, such as selective attention \cite{lin2016neural}, are then taken to aggregate all the sentences in a bag into a single high-quality representation for that bag. 

Distant supervision is used to generate large-scale datasets, and thus some bags will consist of a large number of sentences. These bags can not be split into multiple batches and have to be processed at once during training to construct a single bag-level representation. As a result, MIL is more resource-intensive than fully supervised relation extraction. For example, if there are bags with more than 100 sentences in the training dataset, even by setting the batch size to 1, we require enough hardware memory to pass at least 100 sentences through the sentence encoder to get the bag representation in a step of training. Due to the aforementioned problem, current state-of-the-art methods adopt light-weight and efficient deep neural networks, such as convolutional neural networks (CNN), as the sentence encoder and focus mainly on mitigating the noisy labeling problem.

Deep Transformer neural networks \cite{vaswani2017attention} pre-trained on large corpora \cite{devlin2019bert, radford2018improving, radford2019language} have demonstrated superior capabilities in capturing a contextual semantic representation of words and have achieved state-of-the-art results in many NLP tasks, including supervised relation extraction \cite{soares2019matching, wu2019enriching}. However, these models often have a large number of parameters. They have been shown to capture even better representations as they increase in size \cite{radford2019language, brown2020language}, and thus have significant hardware requirements when training under the MIL framework. To address this issue, we propose a new training method for distantly supervised relation extraction (DSRE). Unlike previous methods that use all the sentences in bags to construct the bag representations, we propose a random instance sampling (RIS) method that limits the number of sentences in a mini-batch by randomly sampling sentences from bags in the mini-batch. Limiting the total number of sentences allows us to leverage deep language representation models such as BERT\cite{devlin2019bert} as the sentence encoder despite limited hardware and produce higher quality sentence representations. However, due to this approach's randomness, using RIS will result in less robust predictions in the inference stage. To mitigate this issue, we train multiple models and then use an ensemble of these models in the inference stage by averaging over prediction probabilities. 
We adopt selective attention as the denoising mechanism and use BERT to encode the relation between entity pairs in sentences.

The contributions of this paper can be summarized as follows:
\begin{itemize}
    \item We propose a new training method for DSRE that relaxes the hardware requirements of MIL by using a random subset of bags in the training phase. This results in a smaller number of sentences in a batch of bags and thus allows us to use larger transformer models as sentence encoders in the distantly supervised setting.
    \item We present two sampling methods for RIS, a baseline that preserves the relative size of the bags after sampling and another method that samples an equal number of sentences from all bags in a mini-batch regardless of their sizes. Our experiments demonstrate the superiority of the latter sampling approach.
    \item We propose the use of an ensemble of different trained models to mitigate the effects of randomness in our new training method. Our experiments demonstrate the effectiveness of this approach.
    \item We use our new training method, coupled with selective attention for bag denoising, to fine-tune BERT on the widely used NYT dataset. Our model achieves an AUC value of 61.4, significantly outperforming previous state-of-the-art methods despite using a simple denoising method. 
\end{itemize}

\section{Related Work}

\citet{mintz2009distant} proposed distant supervision as a way to generate labels for large-scale data for relation extraction automatically. This was done by aligning entities in a knowledge base. However, some of these labels did not match the relation expressed by their corresponding sentences, and thus these noisy labels became the main challenge in DSRE. Subsequent works adopted the MIL paradigm to alleviate the noisy label problem \citep{riedel2010modeling,hoffmann2011knowledge,surdeanu2012multi}, which considered each bag as a sample instead of each sentence. However, these methods used hand-crafted features to encode sentences into vector representations, which limited their performance.

\citet{zeng2015distant} adopted the piecewise convolutional neural network as the sentence encoder and selected only a single sentence in each bag to use as bag-level representation. \citet{lin2016neural} proposed selective attention, which uses a weighted average of sentence representations as the bag-level representation. \citet{liu2017soft} proposed a soft-label method in which bag labels could change depending on the bag-level representation. \citet{qin2018robust} and \citet{feng2018reinforcement} both trained reinforcement learning agents to detect and remove or re-label noisy sentences in bags. \citet{ye2019distant} proposed two novel attention mechanisms, intra-bag attention that considers all relations instead of just the bag's relation to compute the bag-level representation, and inter-bag attention mechanism that aggregates multiple bag representations into a single representation to alleviate the noisy bag problem, i.e., bags with all noisy sentences.

Previous works have incorporated the self-attention mechanism \cite{vaswani2017attention} in their methods \cite{huang2019self, Li2020seg} to address the limitations of piecewise convolutional neural networks (PCNN) in learning sentence representations.

\citet{alt2019fine} extended the OpenAI Generative Pre-trained Transformer (GPT) \cite{radford2018improving} to bag-level relation extraction and used selective attention to compute bag representations. They chose GPT over other transformer models like BERT due to its more reasonable hardware requirements. Our method is similar to \citet{alt2019fine} as we leverage a pre-trained transformer neural network in the distantly supervised setting and use selective attention as the denoising method. However, we use our proposed RIS module during training, which allows us to use BERT as a sentence encoder while requiring much less memory during training.

\section{Proposed Method}
\begin{figure*}
    \centering
    \includegraphics[scale=0.55]{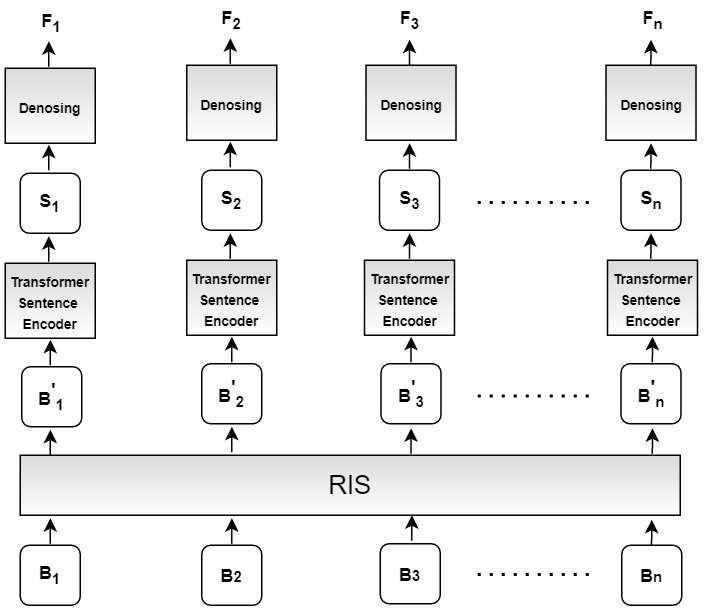}
    \caption{Overview of our proposed sample-based training method}
    \label{fig:Meth}
\end{figure*}

In this section, we present the different steps of the training procedure in our proposed method. In the standard distant supervision setting, the sentences of each bag in a mini-batch are transformed into vector representations using sentence encoders. In the next step, each bag is transformed into a single representation using a denoising method and is then fed into a classification layer. In our proposed training method, a new RIS step is added to the beginning. The new bags computed using RIS are used in the encoding phase instead of the original ones. An essential property of RIS is that it is independent of other steps of training and thus, can be integrated into any other distant supervision method. Overview of our proposed training method is demonstrated in figure \ref{fig:Meth}.

\subsection{Random Instance Sampling}
To address the issue of hardware resource requirements of the MIL framework for distant supervision, we propose a new method called Random Instance Sampling (RIS). Unlike standard supervised relation extraction, batch size in bag-level relation extraction does not control the number of sentences in a mini-batch but only determines the number of bags. Thus, if we are using a sentence encoder with a large number of parameters, such as BERT, it is important to have control over the maximum number of sentences in each step of training to avoid exceeding the available memory. Let \( \text{B} =   \{B_1,..., B_n\} \) denote a mini-batch of bags in a step of training, and \(n\) is the batch size and \( N_{\text{max}}\) denote the maximum number of sentences allowed in a single batch. If the total number of sentences in a mini-batch is less than \( N_{\text{max}}\), the output mini-batch of RIS will be the same as the original one.
\begin{equation}
 N_\text{B} = Sum \{Size(B_1),..., Size(B_n)\}  
\end{equation}
where \(Size(B_i)\) denotes the size of bag \(B_i \) and \( N_\text{B}\) is the total number of sentences in a mini-batch. However, if \( N_\text{B} >N_{\text{max}}  \), due to memory limitations, we will not be able to pass all the sentences through the sentence encoder. Here, we propose two variations of RIS. A baseline method and another approach that improves upon the baseline.
\subsubsection{RIS-baseline}
In the first variation, all bags participate in the sampling regardless of their sizes and the relative bag sizes in a batch are preserved. For all bags, we sample fraction \(f\) of their sentences, where \(f\) is computed by simply dividing \( N_{\text{max}}\) by \(N_\text{B} \). Formally, for each mini-batch B in every step of training, RIS-baseline creates a new mini-batch  \( \text{B}^\prime =   \{B^\prime_1,..., B^\prime_n\} \) as follows:
\begin{equation}
    f = \frac{N_{\text{max}}}{N_\text{B}}
\end{equation}
\begin{equation}
     t_i = Size(B_i) \times f 
\end{equation}
\begin{equation}
    B^\prime_i\ = Sample(B_i, t_i)
\end{equation}

where \( t_i \) denotes how many sentences should be sampled from bag \( B_i\) and \( B^\prime_i\) is the new bag after randomly sampling \(t_i\) sentences from bag \(B_i\)  and mini-batch \(\text{B}^\prime \) will be used instead of mini-batch \(\text{B} \) in the current step of training.
While this is a simple and straightforward approach, it has a significant shortcoming. Distant supervision for relation extraction is based on the assumption that at least one valid sentence exists in each bag. When sampling sentences from small and large bags with equal fractions, the probability that valid sentences in the small bags get removed due to the sampling is higher than that of larger bags, for example, if we have bags with 14 and 2 sentences in a batch of 2 bags with \(N_{\text{max}} = 8 \), then using the RIS-baseline approach we will have bags with 7 and 1 sentences respectively. The probability of removing valid sentences when sampling 1 sentence from a bag of size 2 is higher than when sampling 7 sentences from a bag of size 14. Moreover, the number of valid sentences in bags is usually low and thus, the probability of forming noisy bags (bags without valid sentences) due to sampling is higher for small bags. 

\subsubsection{RIS-equal}
To address the aforementioned issues of RIS-baseline, we propose another approach called RIS-equal. In this approach, we sample an equal number of sentences, denoted by \( N_{\text{sample}}\), from each bag regardless of its size and thus, smaller bags will lose a smaller proportion of their sentences due to sampling than larger bags and bags with sizes less than the aforementioned fixed number will not participate in the sampling. As a result, the overall probability of losing valid sentences will be less than that of RIS-baseline. We set \( N_{\text{sample}}\) to the maximum possible value, which can be calculated by dividing \( N_{\text{max}}\) by the batch size. After the sampling, the total number of sentences in a mini-batch will be less than \( N_{\text{max}}\) if there are bags whose sizes are smaller than \( N_{\text{sample}}\). Let \( N_{\text{diff}}\) be the difference between \( N_{\text{max}}\) and the total number sentences after sampling \( N_{\text{sample}}\) from each bag. If \( N_{\text{diff}}\) is non-zero, we can increase the size of bags whose original sizes before sampling were larger than \( N_{\text{sample}}\) by sampling sentences from their corresponding leftover bags (remaining sentences in bags after sampling \( N_{\text{sample}}\) sentences from them) and adding these sentences to them. We set the number of sentences sampled from each leftover bag to be proportional to the leftover bag size.

Formally, for each mini-batch B in every step of training, RIS-equal creates a new mini-batch  \( \text{B}^\prime =   \{B^\prime_1,..., B^\prime_n\} \) as follows:

\begin{equation}
     N_{\text{sample}} =  \frac{N_{\text{max}}}{n} 
\end{equation}
\begin{equation}
    c_i = \floor*{ \frac{Max\{0, Size(B_i) - N_{\text{sample}}\}}{\sum_j{Max\{0, Size(B_j) - N_{\text{sample}}\}}}}\times N_{\text{diff}} 
\end{equation}
\begin{equation}
     t_i = N_{\text{sample}} + c_i   
\end{equation}
\begin{equation}
    B^\prime_i\ = Sample(B_i, t_i)
\end{equation}
where \(N_sample \) denotes how many sentences we initially sample from each bag, \( t_i \) denotes how many sentences should be sampled from bag \( b_i\) if \( N_{\text{diff}}\) is non-zero and we can sample extra sentences from the leftover bags. Finally, \( B^\prime_i\) is the new bag after randomly sampling \(t_i\) sentences from bag \(B_i\)  and mini-batch \(\text{B}^\prime \) will be used instead of batch \(\text{B} \) in the current step of training.

\begin{table*}
    \centering
    \begin{tabular}{cccccccc}
        \hline
        Method & AUC & P@100 & P@200 & P@300 & P@500 & P@1000 & P@2000 \\
         \hline
         Mintz & 0.106 & 51.8 & 50.0 & 44.8 & 39.6 & 33.6 & 23.4 \\
         PCNN+ATT & 0.336 & 76.3 & 71.1 & 69.4 & 63.9 & 52.7 & 39.1 \\
         PCNN+ATT\_RA+BAG\_ATT & 0.429 & 87.0 & 86.5 & 82.0 & 72.8 & 61.1 & 45.1 \\
         DISTRE & 0.422 & 68.0&67.0&65.3&65.0&60.2&47.9\\
         SeG & 0.51 & 93.0 & 90.0 & 86.0 & 73.5 & 67.0 & 51.6 \\
         \hline
         \hline
         BERT+ATT+RIS-baseline+Ens& 0.608&93.6&
         90.0&87.8&83.0&74.2&58.3\\
         BERT+ATT+RIS-equal+Ens& \textbf{0.614}&\textbf{94.1}&
         \textbf{92.5}&\textbf{90.4}&\textbf{84.6}&\textbf{75.2}&\textbf{58.6}\\
         
         \hline
    \end{tabular}
    \caption{P@N and AUC values of different models}
    \label{tab:patn}
\end{table*}
\subsection{Sentence Encoder}
We follow the approach of \citet{soares2019matching} to encode sentences into relation representations using deep transformers. Similar to \citet{soares2019matching}, we adopt BERT \cite{devlin2019bert} for encoding sentences. After tokenizing the sentences in the dataset, each sentence \(X\) can be represented as a sequence of tokens as follows:
\[ X = [ x_0...x_i...x_j...x_l...x_m...x_n ] \]
where \(x_0 = [CLS] \) and \(x_n = [SEP] \) are special tokens indicating the start and end of the sequence. Sequences \([x_i...x_j] \) and \([x_l...x_m]\) represent tokens for head and tail entities respectively. BERT's output hidden state corresponding to the \([CLS] \) token is used as the sentence representation in task such as sentiment analysis \cite{devlin2019bert}. However, it is not a good representation for relations as it does not make use of the position of entity tokens. \citet{soares2019matching} propose adding special markers before and after head and tail entities as follows:
\[ X^\prime = [ x_0...[H_1]x_i...x_j[H_2]...[T_1]x_l...x_m[T_2]...x_n ] \]
Let \(S_H \) and \(S_T\) denote the output hidden states corresponding to \([H_1] \) and \([T_1] \) respectively. Final vector representation of the relation expressed by each sentence will be computed by concatenating \(S_H \) and \(S_T\) into a single vector \(S\). Formally, each bag \(B = \{X_1, ..., X_m \} \) will become a bag of sentence representations \(\text{S} = \{S_1, ..., S_m \} \) after the encoding stage.

\subsection{Selective Attention}
In order to train a relation classifier in the MIL framework for distant supervision, we need to compute a vector representation for each bag in the dataset. Following \citet{lin2016neural} and \citet{alt2019fine}, we use selective attention to aggregate sentence representations in a bag into a single bag representation. Let \(\text{S} \) denote a bag of sentence representations that mention the same entity pair and \(r\) be the corresponding label provided by distant supervision. Selective attention assigns a weight to each sentence representation in a bag and then computes a weighted average of them to produce the final bag representation. Valid sentences are assigned higher weights, and thus contribute more to the final bag representation whereas noisy sentences will receive lower weights. Final representation of a bag using selective attention can be formulated as follows:
\begin{equation}
    \beta_i = S_i \textbf{r}
\end{equation}
\begin{equation}
    \alpha_i = \frac{exp(\beta_i)}{\sum_j{exp (\beta_j)}}
\end{equation}
\begin{equation}
    F = \sum_i{\alpha_i S_i}
\end{equation}

where \(\textbf{r}\) is a learnable embedding for relation \(r\), \(\beta_i \) is the similarity score between \(r\) and sentence representation \(S_i\), \(\alpha_i\) is the weight assigned to \(S_i\) and \(F\) is the representation of bag \(S\). 

Each bag representation is then fed into a dense layer with softmax activation to compute the probability distribution over all the relations.
\begin{equation}
    d = \textbf{W} F + \textbf{b}
\end{equation}
\begin{equation}
    P(\textbf{r}|S, \theta) = softmax(d)
\end{equation}
where \textbf{W} and \textbf{b} are the learnable parameters of the Dense layer, \(\theta\) denotes the model's parameters and \(P(r|S, \theta) \) is the probability distribution over relation labels.

We formulate the objective function of the training as follows:
\begin{equation}
    \mathcal{J}_D = \sum^{|D|}_i{P(r_i|\text{S}_i, \theta)}
\end{equation}

Where \(|D| \) denotes the number of bags in the training set.

\subsection{Ensemble Modeling for Prediction}
Using RIS for training will cause two main issues:
\begin{itemize}
    \item During training, we apply RIS to each mini-batch at each training step. Thus some valid sentences may not participate in the construction of their corresponding bag's representation. This negatively affects the model's convergence in training and thus, impacts the model's performance in the evaluation stage.
    \item Each time we train the model, it is practically trained on a slightly different dataset due to RIS. Thus, the performance will vary each time we train the model from scratch.
\end{itemize}
\begin{table}[]
    \centering
    \begin{tabular}{|cc|}
    \hline
         \textbf{Parameter}&\textbf{Value}  \\
         \hline
      
         Optimizer&Adam \\
         Learning Rate & 5e-6\\
         Batch Size& 12\\
         Max sentence in Batch& 36\\
        \hline
    \end{tabular}
    \caption{Hyper-parameters used in our experiments}
    \label{tab:hparam}
\end{table}

\begin{table*}
    \centering
    \begin{tabular}{cccccccc}
        \hline
        Method & AUC & P@100 & P@200 & P@300 & P@500 & P@1000 & P@2000 \\
         \hline
         BERT+ATT+RIS-baseline & 0.544 & 89.1 & 85.1 & 82.7 & 78.0 & 69.5 & 54.6 \\
         + Ensemble (n=2) & 0.578 & 91.7 & 89.0 & 85.1 & 80.5 & 72.1 & 56.6 \\
         + Ensemble (n=3) & 0.596 & 93.2 & 90.5 & 86.8 & 81.8 & 73.9 & 57.5 \\
         + Ensemble (n=4) & 0.604 & 93.2 &89.7&87.0&82.4&74.0&58.1\\
         + Ensemble (n=5) & 0.608&93.6&
         90.0&87.8&83.0&74.2&58.3 \\
         \hline
         \hline
          BERT+ATT+RIS-equal & 0.555 & 89.7 & 86.1 & 83.9 & 80.1 & 70.5 & 55.0 \\
         + Ensemble (n=2) & 0.581 & 91.5 & 89.3 & 87.8 & 81.9 & 72.1 & 56.5 \\
         + Ensemble (n=3) & 0.601 & 93.9 & 91.4 & 89.7 & 84.1 & 73.8 & 57.6 \\
         + Ensemble (n=4) & 0.606 & 95.0&91.9&89.0&83.4&74.2&58.2\\
         + Ensemble (n=5) & 0.614&94.1&
         92.5&90.4&84.6&75.2&58.6 \\
         
         \hline
    \end{tabular}
    \caption{P@N and AUC values for different number of trained models used for ensemble}
    \label{tab:ens}
\end{table*}

Due to the issues mentioned above, we propose training several models and using an ensemble of those models for evaluation. An ensemble method mitigates the first issue because specific valid sentences that are ignored during training for one model may take part in the training of other models. For the second issue, using an ensemble of multiple models for evaluation will reduce the predictions' variance.

Let \(\text{M} = \{m^1,...,m^n\} \) denote a list of predictions from multiple trained models, corresponding to a bag in the test set, and \(n\) is the number of models. Let \(m^i = \{p^i_1,...,p^i_l\}\) denote the scores predicted by model \(m^i\), and \(l\) is the number of relations. Then, the final score predicted by the ensemble of \(n\) models for each relation is computed by taking the unweighted average of the scores predicted by all the models for that relation.

\begin{equation}
    p^\text{Ens}_j = \frac{\sum^n_{i=1}{p^i_j}}{n}
\end{equation}
where \(p^\text{Ens}_j\) denotes the score predicted by the ensemble for relation \(j\) for a bag in the test set.

\section{Experiments}
\subsection{Dataset}
We evaluate our model on the widely used NYT dataset \cite{riedel2010modeling} which was generated by aligning freebase with the New York Times corpus. Articles from 2005 to 2006 were used for the training set, and articles from 2007 were used for the test set. There are 53 distinct relation types in the dataset, including the special NA relation which indicates the lack of semantic relation between entity pairs. The training set contains 570K sentences, and the test set contains 170K sentences.

\subsection{Evaluation Metrics}

Following previous works, we use area under the curve (AUC), Precision@N (P@N) and precision-recall (PR) curves to evaluate our model on the held-out test set of the NYT dataset.

\subsection{Implementation Detail}

We extend the OpenNRE framework \cite{han-etal-2019-opennre} to implement our model. We use BERT\textsubscript{Large} pre-trained model\footnote{https://github.com/google-research/bert} released by \citet{devlin2019bert} to initialize BERT. It has 24 encoder layers, 16 attention heads, and 1024 hidden state size. For hyper-parameter tuning, we use 20 percent of the training set as validation set and selected hyper-parameters that result in the best AUC value on the validation set. Table \ref{tab:hparam} shows the hyper-parameters used in our experiments. We train our model on the full training set for 2 epochs using these hyper-parameters. We use an ensemble of 5 models for prediction in the test stage, and all reported results are the average of 5 different runs. All our experiments were conducted using a single Tesla T4 GPU.

\begin{table*}
    \centering
    \begin{tabular}{cccccccc}
        \hline
        \( N_{\text{max}}\) & AUC & P@100 & P@200 & P@300 & P@500 & P@1000 & P@2000 \\
         \hline
         24 & 0.598  & 92.1 & 88.6 & 86.2 & 80.3 & 73.4 & 57.9 \\
         36 & 0.608  &93.6& 90.0&87.8&83.0&74.2&58.3 \\
         48 & 0.594 & 91.9&88.4&85.8&80.0&72.9&57.6\\
    
         \hline
        
    \end{tabular}
    \caption{P@N and AUC values for different values of \( N_{\text{max}}\) when using RIS-baseline}
    \label{tab:mx1}
\end{table*}
\begin{table*}
    \centering
    \begin{tabular}{cccccccc}
        \hline
        \( N_{\text{max}}\) & AUC & P@100 & P@200 & P@300 & P@500 & P@1000 & P@2000 \\
         \hline
         24 & 0.603   & 93.1 & 89.0 & 86.6 & 81.5 & 73.9 & 58.4 \\
         36 & 0.614 &94.1&
         92.5&90.4&84.6&75.2&58.6 \\
         48 & 0.602&94.7&
         89.9&87.5&82.3&74.2&58.0\\
    
         \hline
        
    \end{tabular}
    \caption{P@N and AUC values for different values of \( N_{\text{max}}\) when using RIS-equal}
    \label{tab:mx2}
\end{table*}

\subsection{Baselines}
We compare our model with the following baselines:
\begin{itemize}
    \item Mintz \cite{mintz2009distant} is the original distant supervision model that uses hand-crafted features and a logistic regression classifier.
    \item PCNN+ATT \cite{lin2016neural} adopts the PCNN for sentence encoding and uses the selective attention mechanism to compute bag representations.
    \item PCNN+ATT\_RA+BAG\_ATT \cite{ye2019distant} adopts the PCNN as sentence encoder and uses inter-bag and intra-bag attention mechanisms for denoising.
    \item DISTRE \cite{alt2019fine} adopts the GPT as sentence encoder and performs bag-level relation extraction using selective attention.
    \item SeG \cite{Li2020seg} adopts a self-attention enhanced PCNN along with entity-aware embeddings to represent sentences and uses a selective gate mechanism for denoising.
\end{itemize}

\begin{figure}
    \centering
    \includegraphics[scale=0.55]{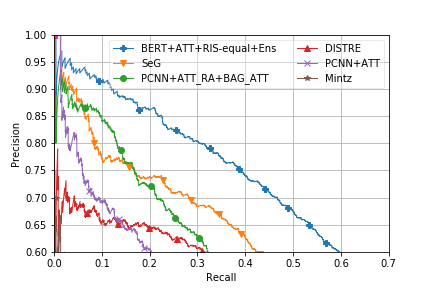}
    \caption{PR curves comparison of our proposed method and different baselines}
    \label{fig:prec-rec}
\end{figure}

\subsection{Evaluation Results}
 Table \ref{tab:patn} shows AUC and P@N values of our proposed method and different baselines. The last two rows of the table show the results of our model when used with the two proposed sampling methods. We used an ensemble of 5 different trained models in our experiments. We observe that RIS-equal achieves the highest AUC value and significantly outperforms the previous state-of-the-art method by 0.104. Our method also outperforms all other baselines in P@N for all the values of N up to 2000 and in almost all recall levels. Figure \ref{fig:prec-rec} shows the PR curves of RIS-equal and baseline models.
 
\begin{figure}
    \centering
    \includegraphics[scale=0.55]{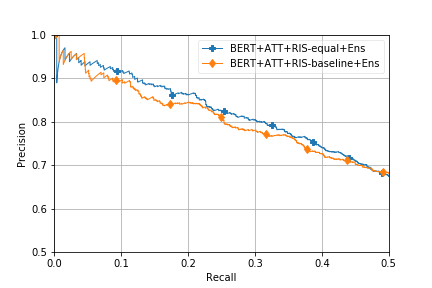}
    \caption{PR curves comparison of our model when used with our two proposed sampling methods}
    \label{fig:eq_vs_baseline}
\end{figure}

 We also observe that RIS-equal outperforms RIS-baseline in both AUC and P@N, which proves the effectiveness of sampling larger proportions from smaller bags. Figure \ref{fig:eq_vs_baseline} shows the PR curves of RIS-equal and RIS-baseline.
 
 These results demonstrate that we can achieve state-of-the-art performance even when using a subset of bags in training in the distantly supervised setting. 
 
 Despite the similarities of our method with DISTRE, our evaluation results show a significant gap in the performance of the two methods. This shows the difference between the quality of language representations produced by GPT and BERT\textsubscript{Large}. While BERT\textsubscript{large} has three times the number of parameters of GPT, our model requires much less memory for training compared with DISTRE, which shows the effectiveness of using RIS in training.
 
Our method also achieves much better results than SeG and PCNN+ATT\_RA+BAG\_ATT despite using a less effective denoising method. This indicates the importance of using better sentence representations compared with using better denoising methods.

\subsection{Effectiveness of Ensemble Modeling}
In this section, we conduct extensive experiments to show the effectiveness of using Ensembles for alleviating the randomness of RIS. We report P@N and AUC values for different numbers of trained models used for ensemble in Table \ref{tab:ens}. BERT+ATT+RIS-baseline and BERT+ATT+RIS-equal indicate the two methods without using ensemble modeling. As shown in the table, Increasing the number of trained models results in increased AUC and P@N for both RIS-baseline and RIS-equal. Overall, using more trained models for ensemble results in more training and evaluation time and thus, the choice of the number of models used for the ensemble is a trade-off between performance and speed.

\subsection{Effect of Maximum Number of Sentences in Batch}
We conducted experiments with different values of \( N_{\text{max}}\). We tested three different values of 24, 36, and 48. Using a single GPU with 16 GBs of memory, 48 was the maximum value of \( N_{\text{max}}\) we could set. The results of the experiments for both BERT+ATT+RIS-equal and BERT+ATT+RIS-baseline are demonstrated in Table \ref{tab:mx2} and Table \ref{tab:mx1} respectively. We expected our models to perform better for higher values of \( N_{\text{max}}\) as sampling a higher number of sentences from bags would result in a lower probability of losing valid sentences from bags. However, we achieved the best results when setting \( N_{\text{max}}\) to 36 for both sampling methods. This could be attributed to selective attention because when the number of sentences in a bag is smaller, the effect of the valid sentences in the weighted sum increases. Thus, when we set \( N_{\text{max}}\) to a small value, the resulting bags become smaller, and better bag representations could be computed. However, the probability of losing valid sentences increases, which negatively affects the quality of bag representations. The opposite holds for high values of \( N_{\text{max}}\).

\section{Conclusion}
In this paper, we proposed a new sample-based training method for distantly supervised relation extraction that reduces the hardware requirements of the multiple instance learning paradigm by randomly sampling sentences from bags in a batch. We then alleviated the issues raised by this randomness by using an ensemble of multiple trained models. The reduced hardware requirements allowed us to leverage a pre-trained BERT model for relation extraction in the distantly supervised setting. Experimental results on the widely used NYT dataset demonstrated that our method significantly outperforms current state-of-the-art methods in terms of both AUC and P@N values.



\bibliography{anthology,eacl2021}
\bibliographystyle{acl_natbib}

\end{document}